\documentclass[letterpaper]{article} 
\usepackage[preprint]{aaai2027}  
\usepackage[hyphens]{url}  
\usepackage{graphicx} 
\usepackage{natbib}  
\usepackage{caption} 
\usepackage{algorithm}
\usepackage{algorithmic}

\usepackage{newfloat}
\usepackage{listings}
\DeclareCaptionStyle{ruled}{labelfont=normalfont,labelsep=colon,strut=off} 
\floatstyle{ruled}
\newfloat{listing}{tb}{lst}{}
\floatname{listing}{Listing}

\usepackage{booktabs}

\title{Shortcut Mitigation via Spurious-Positive Samples}

\author {
    Phuong Quynh Le\textsuperscript{\rm 1}\corresponding,
    Jörg Schlötterer\textsuperscript{\rm 1},
    Sari Sadiya\textsuperscript{\rm 2},
    Gemma Roig\textsuperscript{\rm 2},
    Christin Seifert\textsuperscript{\rm 1}
}
\affiliations {
    \textsuperscript{\rm 1}Marburg University, Germany\\
    \textsuperscript{\rm 2}Goethe University Frankfurt, Germany\\
}

\usepackage[utf8]{inputenc} 
\usepackage[T1]{fontenc}    
\usepackage{booktabs}       
\usepackage{amsfonts}       
\usepackage{nicefrac}       
\usepackage{microtype}      

\usepackage{xcolor, colortbl}
\usepackage{amsmath}
\usepackage{amssymb}
\usepackage{pifont}
\usepackage{caption}
\usepackage{subcaption}
\usepackage{multirow}
\usepackage{color}
\usepackage{dsfont}
\usepackage{enumitem}
\usepackage{afterpage} 
\usepackage{array,booktabs}

\usepackage{xr}

\usepackage{xspace}
\newcommand{\modelname}{\textsc{SCORE}\xspace}

\usepackage{xspace}

\usepackage{threeparttable}

\begin{document}

\maketitle


\begin{abstract}
Shortcut mitigation strategies commonly rely on training data annotations, group-balanced held-out data or the presence of all groups, i.e., all combinations of (spurious) attributes and classes, in the training data. However, these requirements are rarely met in practice.
We instead propose a method for targeted model analysis to identify a small set of instances in which the model relies on spurious attributes.
Using that set and following ``this feature should not be used for prediction'' reasoning, we identify highly relevant neurons in an intermediate layer and regularize their impact.
This ensures that models learn to depend on informative features rather than being right for the wrong reasons, thereby improving robustness without requiring additional balanced held-out data or annotations.
\end{abstract}
\section{Introduction}
\label{sec:intro}
Neural networks have driven tremendous progress in image classification. However, diagnosing pneumonia based on hospital identifiers visible in CT scans~\cite{Zech:PLOS:2018} is not a reliable form of recognition. This phenomenon, known as spurious correlation learning (also referred to as shortcut learning), occurs when models exploit features that are strongly correlated with the target class but lack causal relevance. 
Models that rely on spurious correlations often have strong overall predictive performance, but their reasoning is flawed -- they are ``right for the wrong reason''.
Prior research has shown this phenomenon in both synthetic and realistic settings: models tend to rely on image textures for object classification~\cite{Geirhos:2018:ICLR}, background features for bird types classification~\cite{Sagawa:2020a:ICLR}, gender features for hair color prediction~\cite{Liu:2015:ICCV}, and color patches for skin cancer diagnosis~\cite{Nauta:2021:diagnostics}. 

\begin{figure}[t]
\begin{center}
\includegraphics[width=\columnwidth, trim={20cm 16cm 20cm 6.5cm},clip]{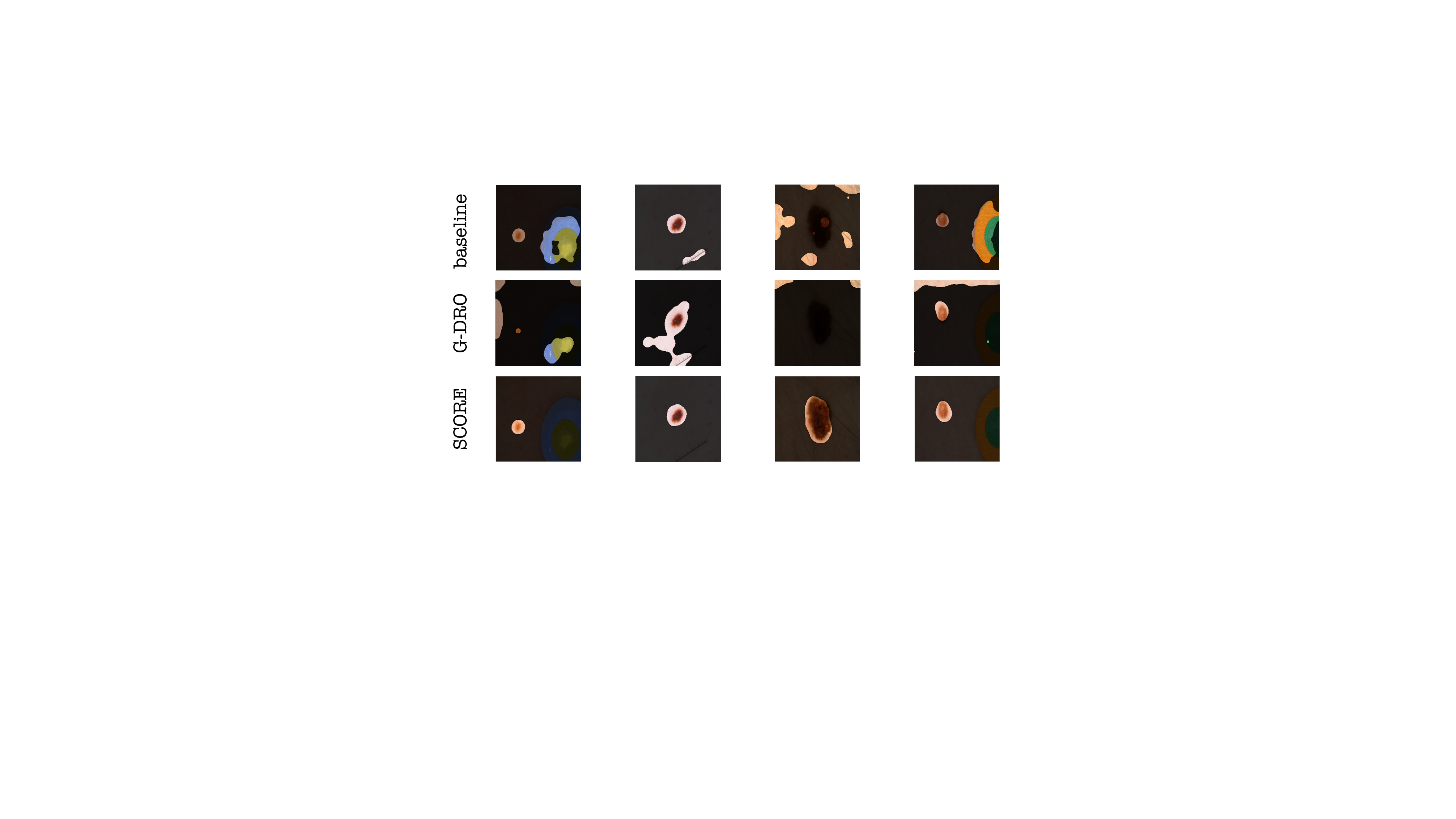}
\end{center}
\caption{LRP heatmaps for samples from the baseline model (top), model trained by GDRO (middle) and after fine-tuning using SCORE (bottom). The results show that our proposed regularization effectively eliminates the reliance on spurious features (color patches, rulers, hairs), whereas GDRO, despite using full annotations, fails to eliminate this reliance.}
\label{fig:preview}
\end{figure}

\begin{figure*}[tb]
        \centering
    \includegraphics[width=\linewidth]{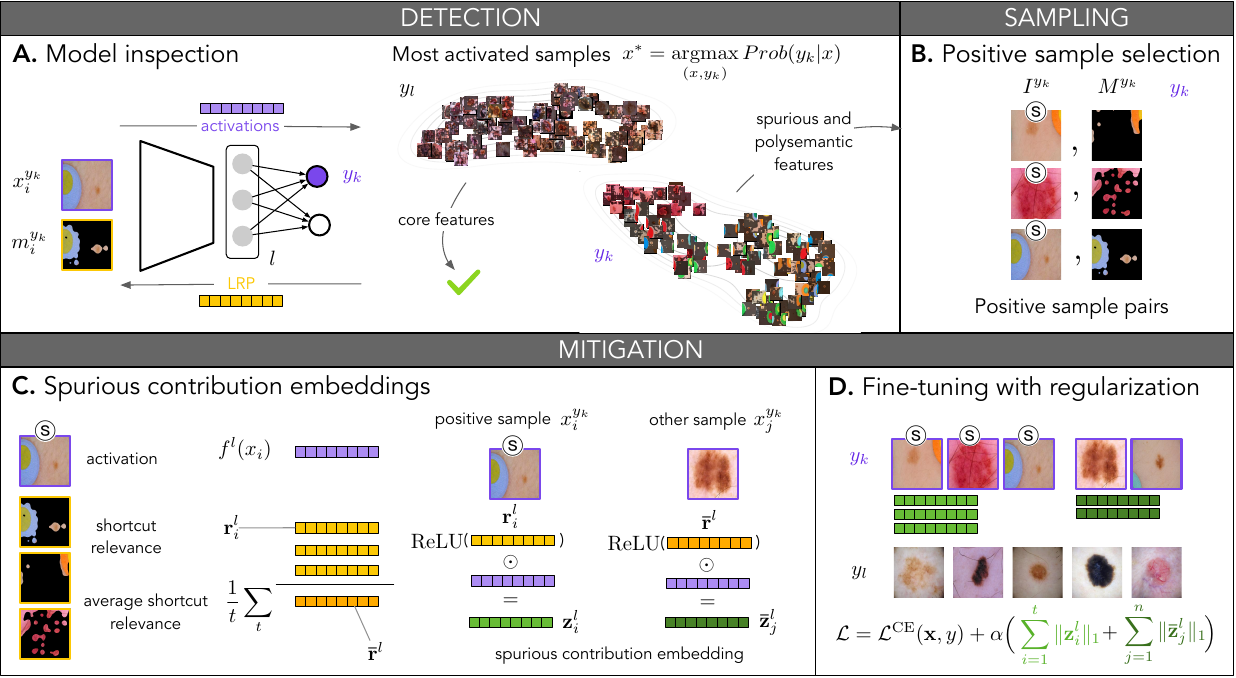}
    \caption{Overview of \modelname. \textbf{A.} For each instance, we compute activations and LRP masks, then detect potential spurious features based on the most relevant features. \textbf{B.} Among the most activated samples per class, we select \textit{spurious-positive samples} using the masked images. 
    \textbf{C.} For these samples, we extract activation vectors from the original inputs and shortcut relevance vectors from the masked images. We apply ReLU to the shortcut relevance vectors and combine them element-wise with the activations to obtain spurious contribution embeddings. An average shortcut relevance vector is further used to approximate spurious contribution embeddings for other samples of the spurious class. 
    \textbf{D.} We fine-tune the model on class-balanced batches, applying regularization terms to sparsify spurious contribution embeddings.}
    \label{fig:approach-overview}
\end{figure*}

Formally, let $\mathcal{X}$ be the input space, $\mathcal{Y}$ the target space, and $\mathcal{S}$ a space of (\emph{spurious}) attributes. Each example is a triple $(x,y,s)$ with $x\in\mathcal{X}$, $y\in\mathcal{Y}$, and $s\in\mathcal{S}$. Training and test samples are drawn from distributions $P_{\mathrm{train}}$ and $P_{\mathrm{test}}$ respectively, over $(\mathcal{X},\mathcal{Y},\mathcal{S})$. Every example belongs to a \emph{group}
\[
g=(y,s)\in\mathcal{G}:=\mathcal{Y}\times\mathcal{S}.
\]
Here \(\mathcal{G}\) is the set of possible combinations of attributes $s$ and targets $y$. 
Models tend to learn spurious correlations when the data distribution \(\mathcal{G}\) is highly imbalanced, particularly between a specific target-spurious attribute group and other groups. 
This is especially the case when the spurious attribute is easy to learn, as it accelerates the learning process and causes the model to converge quickly. As a result, the models often rely heavily on this spurious attribute while neglecting other core informative features, leading to poor generalization.

To mitigate spurious correlation learning, most approaches explicitly address the imbalanced data distributions across groups by incorporating group labels $g$ during training~\cite{Sagawa:2020a:ICLR,Srivastava:2020:ICML,Rieger:2020:ICML} or in the validation set for partial retraining and hyperparameter tuning~\cite{Kirichenko:2023:ICLR,Labonte:NeurIPS:2024}. The main disadvantage of this method is that it requires explicit group labels. Other approaches first infer group annotations implicitly by hard and easy cases, and then upweight the hard cases~\cite{Liu:2021:ICML,Nam:2020:NeurIPS,ZhangMichael:2022:ICML}. 
In general, these approaches assume that all possible combinations of pairs $(y, s)$, i.e., the complete set of groups $\mathcal{G}$, exist in the training data.
Clearly, group annotations for all training instances are unrealistic, not only because of the high annotation effort, but also because groups are not known a priori. Only upon model analyses after training, reliance on spurious correlations and the corresponding groups becomes apparent. While collecting a small group-balanced validation set for partial retraining seems a manageable effort, synthetic benchmarks obscure real-world challenges. 

We argue that neither fully annotated training data nor group-balanced validation sets are available in reality, but annotation of a few instances for which the model relies on spurious features is a reasonable effort. In particular, since spurious correlation learning is typically discovered through such instances during model analysis, and their annotation introduces little overhead.
We propose a framework, that does not require group labels $g$, nor that all groups are present in the training dataset.
For targeted model analysis, we first use layer-wise relevance propagation (LRP) to detect potential spurious correlation learning and to identify a small set of \emph{spurious-positive samples} (i.e., instances where the model relies on a shortcut feature in its prediction), following ``this feature should not be used'' reasoning. 
Notably, the required human annotation effort in this targeted analysis is comparable to that typically involved in model explanation and interpretability analyses. 
To mitigate the model's reliance on spurious features, we propose a novel regularization \modelname (Spurious COntribution REgularization) that penalizes the activation of neurons that are highly relevant to spurious features in the identified spurious-positive samples. 

We summarize our contributions as follows: (i) We demonstrate the effectiveness of using LRP to detect spurious correlations, requiring only minimal human effort.
(ii) We propose a regularization mechanism for mitigating spurious correlations that does not require held-out data or group-balanced annotations. 
(iii) Unlike existing methods, our approach remains effective even when data from certain groups is completely missing.

\begin{figure*}[t]
        \centering
\includegraphics[width=\linewidth]{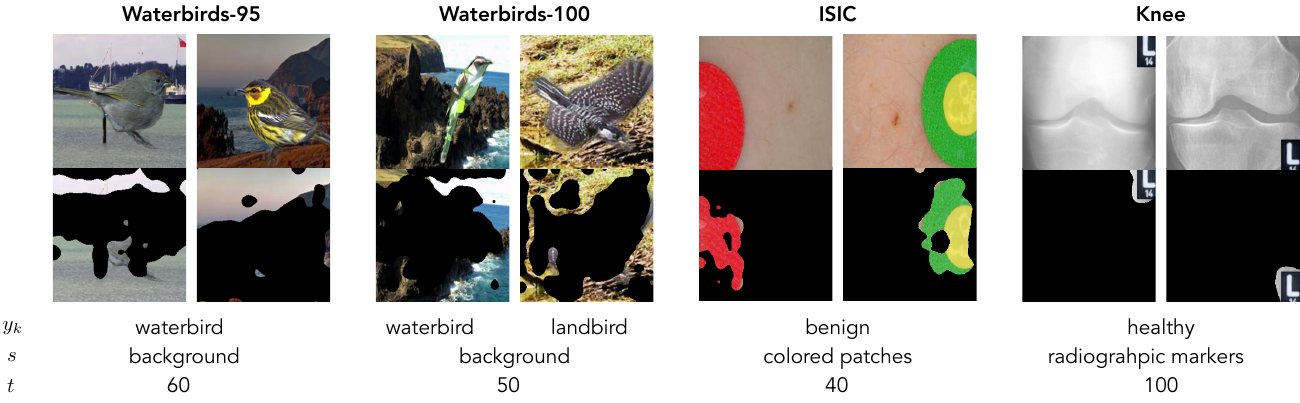} 
    \caption{Selected LRP masks for shortcut-positive samples. For each dataset, we show two example pairs of masked and original image $(m_i,x_i)$ used for regularization, the spurious attribute $s$, and the class which $s$ is spurious-positive for.  From 100 analyzed samples, we retain $t$ samples that clearly highlight spurious features; $t$ being specific for each dataset.}
    \label{fig:lrp-mask}
\end{figure*}

\section{Related Work}
\label{sec:related-work}

\subsubsection{Shortcut Mitigation using Group Samples.} 
Prior work on mitigating spurious correlations largely assumes access to group information or mechanisms to identify bias-aligned (i.e., spuriously correlated) samples. These methods broadly fall into two categories. 
The first category leverages group annotations, either explicitly or implicitly, to optimize robust objectives. GroupDRO~\cite{Sagawa:2020b:ICML} minimizes the worst-group loss using full group labels, while MRM~\cite{Zhang:2021:ICML} constructs group-aware sub-tasks to learn robust subnetworks. To reduce annotation requirements, DFR~\cite{Kirichenko:2023:ICLR}, AFR~\cite{Qiu:2023:ICML}, and SELF~\cite{Labonte:NeurIPS:2024} fine-tune models using a small set of group- or class-balanced samples. However, these approaches typically rely on a held-out, relatively clean validation set and assume knowledge of the group distribution $\mathcal{G}$, which may be unavailable in practice. 
The second category identifies bias-aligned and bias-conflicting samples via training dynamics. LfF~\cite{Nam:2020:NeurIPS}, JTT~\cite{Liu:2021:ICML}, and DEDIER~\cite{Tiwari:2024:WACV} emphasize high-loss or misclassified samples under the assumption that they are not correlated with spurious features (bias-conflicting). TAB~\cite{Espinosa:2024:ECCV} further clusters samples based on loss trajectories to enable group-balanced retraining. Related extensions include CnC~\cite{ZhangMichael:2022:ICML} and DCWP~\cite{Park:2023:CVPR}, which incorporate contrastive learning or specialized subnetworks. These methods are sensitive to the quality of the initial ERM model and often require careful early stopping. PruSC~\cite{Le:TMLR:2025} mitigates this issue by clustering samples in intermediate representation space without relying on explicit misclassifications.

\subsubsection{Explain-then-Mitigate.} 
Another line of work follows an explain-then-mitigate paradigm, where spurious features are first identified using explainability methods and subsequently suppressed during training. DISC~\cite{Wu:2023:ICML} uses a predefined concept bank to identify spurious concepts and rebalance the dataset accordingly. CDEP~\cite{Rieger:2020:ICML}, RRR~\cite{Ross:2017}, and related attribution-based methods~\cite{Chefer:2022:NeurIPS} regularize models by penalizing explanations over undesired regions, typically requiring concept annotations or pixel-level masks. 
More recently, multimodal large language models have been used to automatically annotate visual concepts and separate core from spurious features~\cite{Petryk:2022:CVPR,Chakraborty:2024:NeurIPS,Kuhn:2025:ICCV}. ASM~\cite{Kuhn:2025:ICCV}, for example, names concept clusters derived from intermediate ViT representations and mitigates shortcut learning by suppressing patches associated with spurious concepts, but relies on held-out data for concept discovery. 

Overall, existing approaches largely depend on extensive annotations, clean validation data, or complete group information, limiting their applicability in real-world settings.

\section{Model Inspection and Sample Identification}
Figure~\ref{fig:approach-overview} provides an overview of our \modelname framework to identify and suppress internal model representations that encode spurious features while preserving class-discriminative information. 
Since spurious features are task-dependent and generally unknown \emph{a priori}, we first in this section identify spurious samples semi-automatically by computing LRP masks and visualizing the evidence for the most highly activated samples per class. We refer to samples whose predictions rely on spurious features as \textbf{spurious-positive samples}.

\subsubsection{Layer-wise Relevance Propagation.}
\label{sec:background}

Consider a network $f$ composed of $L$ layers, mapping an input $x$ to an output
\begin{equation}
    f(x) = f^{L} \circ \cdots \circ f^{l} \circ f^{l-1} \circ \cdots \circ f^{1}(x).
    \label{eq:lrp_network}
\end{equation}
Layer-wise Relevance Propagation (LRP)~\cite{Bach:PLOS:2015}, assigns an initial relevance score to an output neuron and propagates this relevance backward through the network to quantify the contribution of each neuron, and ultimately each input dimension, to the prediction.

For a given layer, let $a_{mn}$ denote the pre-activation contribution from neuron $m$ in layer $l-1$ to neuron $n$ in layer $l$, and let $a_n = \sum_m a_{mn}$ be the aggregated pre-activation of neuron $n$. LRP redistributes the relevance $R^l_n$ received at neuron $n$ in layer $l$ to neurons in the preceding layer proportionally to their relative contributions, according to
\begin{equation}
    R^{(l-1,l)}_{m \leftarrow n} = \frac{a_{mn}}{a_n} \, R^l_n .
    \label{eq:lrp_message}
\end{equation}
The relevance message $R^{(l-1,l)}_{m \leftarrow n}$ thus measures the contribution of neuron $m$ in layer $l-1$ to the activation of neuron $n$ in layer $l$. We use $\epsilon$-LRP for signed relevance to target class.

\subsubsection{Spurious-positive Sample Selection.}
LRP assigns relevance scores that quantify the proportional contribution of each neuron to a target output. When computed with respect to input features and visualized as a heatmap, these scores highlight the input regions that are most influential for the model’s decision. We hypothesize that the features that are critical for predicting a given class are consistently present in the samples for which the model is most confident, i.e., samples with the highest activations (logits). Consequently, if the model relies on shortcut features, these shortcuts should manifest among the most relevant features in the most activated samples associated with the target class.
Fig. \ref{fig:approach-overview},A shows an example when we visualize the 100 most activated samples for each of the classes benign ($y_k$) and malignant ($y_l$) in the ISIC dataset. For LRP heatmaps of class malignant, we do not find any potential spurious features. For LRP heatmaps of class benign, while there are still samples highlighting core features of lesions, we observe a large proportion recognizing the ``color patches'' as relevant features, which shows possible shortcut learning. 

We characterize a given class $y_k$ using the top-$n_{\text{ref}}$ samples with the highest prediction scores under the baseline model $f$. For each of these samples, we compute input-level relevance masks by propagating relevance scores to the input using LRP. This yields visualizations of the image regions that are most influential for predicting class $y_k$.
Based on visual inspection of the resulting masked images, we identify potential spurious correlations learned by the model. Once such features are detected, we manually select a subset of $t$ samples that prominently exhibit them and denote these as \textbf{spurious-positive samples}. We collect these samples in the \emph{positive sample set} $I^{y_k} = \{ i_1, \dots, i_t \}$ and their corresponding masked images in the \emph{positive mask set} $M^{y_k} = \{ m_1, \dots, m_t \}$, where the masks highlight spurious features associated with class $y_k$.

\begin{figure}[t]
\begin{center}
\includegraphics[width=\columnwidth]{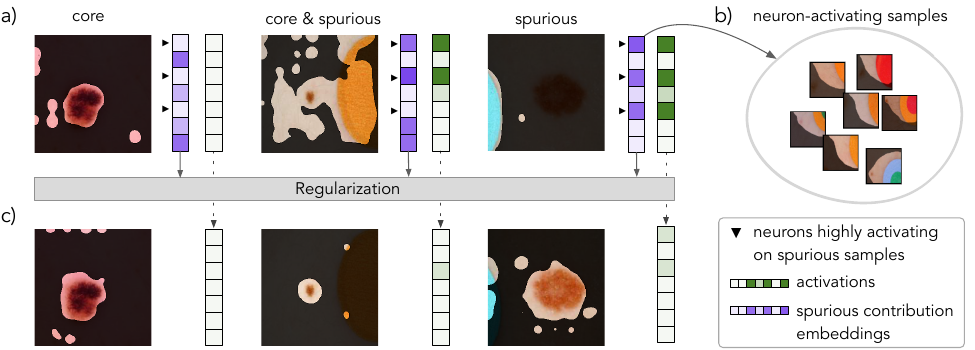}
\end{center}
\caption{a) LRP activations show that the model uses spurious features. b) Most strongly activated neurons for spurious-positive samples encode spurious semantic concepts. (c) \modelname penalizes reliance on spurious-activating neurons, forcing the model to learn informative features.}
\label{fig:teaser}
\end{figure}
\subsubsection{Observation.}
Unlike methods that infer spuriousness indirectly, such as high loss, missclassified cases or clustering structure, our inspection stage names the spurious features explicitly, through human inspection of LRP evidence on the most confident samples. A model can therefore be flagged as flawed even when its accuracy is high, and a class may legitimately carry no shortcut, as for malignant in ISIC. 
We first present results of shortcut inspection and LRP mask visualization for all datasets in Fig. \ref{fig:lrp-mask}, exemplifying the identified spurious regions and the corresponding positive sample sets.

Beyond input-level evidence, we analyze how spurious regions are encoded in the network’s intermediate layers. When passing the masked inputs $M^{y_k}$ through the network, we observe a small, consistent subset of neurons in layer~$l$ activated. Following the semantic neuron analysis of Dreyer et al.~\cite{Dreyer:CVPR:2024}, we find that the most strongly activated neurons correspond to spurious semantic concepts (see Fig. \ref{fig:teaser},b), in line with recent findings that spurious features occupy separable neuronal subspaces~\cite{dong:ICML:2025,Le:2025:xai}. This motivates suppressing these neurons directly, and averaging their relevance over the spurious-positive samples in order to extend the suppression to samples we did not annotate. Penalizing in representation space removes the need for the per-sample pixel masks required by RRR~\cite{Ross:2017} and CDEP~\cite{Rieger:2020:ICML}. 
Analyses of LRP-based shortcut detection, the effect of varying $n_{\text{ref}}$, are provided in Suppl. \ref{suppl:sec:shortcut-detection}.

\section{Shortcut Mitigation}
In the direction of motivating analysis, we compute \textbf{spurious contribution embeddings}, which capture neuron-level contributions associated with spurious input features. To reduce the model’s dependence on such features, we apply $\ell_1$ regularization to the spurious contribution embeddings, biasing the model toward sparse representations of spurious features. We then fine-tune the network using a joint objective that enforces both class discrimination and reduced reliance on spurious features

\begin{table*}[t!bp]
    \centering
    \setlength{\tabcolsep}{4pt}
    \begin{threeparttable}
     \begin{tabular}{lcccccccccc}
    \toprule
         & \multirow{2}{0.8cm}{group label} & \multirow{2}{0.8cm}{masked annot.} & \multicolumn{2}{c}{Waterbirds-95} & \multicolumn{2}{c}{Waterbirds-100} &  \multicolumn{2}{c}{ISIC} & \multicolumn{2}{c}{Knee}\\
         & &  & AVG & WGA$\uparrow$ & AVG & WGA$\uparrow$ & AVG & WGA$\uparrow$ & AVG & WGA$\uparrow$\\
         \midrule
       ResNet50  &\ding{55} &\ding{55} & 90.1{\scriptsize $\pm$0.3} & 75.3{\scriptsize $\pm$0.6} & 74.9{\scriptsize $\pm$0.8} & 36.5{\scriptsize $\pm$4.3} & 86.2{\scriptsize $\pm$0.2} & 34.4{\scriptsize $\pm$0.8} & 65.9{\scriptsize $\pm$0.4} & 35.3{\scriptsize $\pm$1.2}\\
       \midrule
       GroupDRO &\ding{51} &\ding{55}& 92.0{\scriptsize $\pm$0.6} &  \textbf{89.9}{\scriptsize $\pm$0.6} & 80.4{\scriptsize $\pm$0.8}  & 56.7{\scriptsize $\pm$1.3} &  87.7{\scriptsize $\pm$0.8} &  59.0{\scriptsize $\pm$0.4} & 72.8{\scriptsize $\pm$0.4} & 34.3{\scriptsize $\pm$0.6}\\
       DFR$_{\mathrm{Tr}}$  &\ding{51} &\ding{55} & 87.5{\scriptsize $\pm$1.2} & 74.6{\scriptsize $\pm$2.9} & 80.3{\scriptsize $\pm$5.5} & 34.7{\scriptsize $\pm$4.3} & 74.1{\scriptsize $\pm$2.6} & 48.8{\scriptsize $\pm$6.0} & 80.8{\scriptsize $\pm$3.5} & 26.0{\scriptsize $\pm$5.6}\\
       \midrule
       CnC  &\ding{55} &\ding{55}& 90.9 {\scriptsize $\pm$ 0.1} & 88.5 {\scriptsize $\pm$ 0.3} &  &  &    &     &  79.1{\scriptsize $\pm$ } &  62.0{\scriptsize $\pm$ } \\
       JTT  &\ding{55} &\ding{55}& 89.3{\scriptsize $\pm$0.7} & 83.8{\scriptsize $\pm$1.2} & 78.5{\scriptsize $\pm$0.9} & 25.4{\scriptsize $\pm$0.3} & 77.8{\scriptsize $\pm$2.3} & 20.1{\scriptsize $\pm$3.4} & 60.4{\scriptsize $\pm$0.6} & 38.0{\scriptsize $\pm$2.0}\\
       PruSC  &\ding{55} &\ding{55}& 90.3{\scriptsize $\pm$2.2} & 79.8{\scriptsize $\pm$0.8} & 90.6{\scriptsize $\pm$0.6} & 67.1{\scriptsize $\pm$0.4} & 86.1{\scriptsize $\pm$0.4} & 75.1{\scriptsize $\pm$1.7} &  77.3{\scriptsize $\pm$2.3} &  67.0{\scriptsize $\pm$0.8} \\
       \midrule
       CDEP  &\ding{55} &\text{fullset}&  89.6{\scriptsize $\pm$ } &  69.3{\scriptsize $\pm$ } & 61.0 &  27.4  &  67.8{\scriptsize $\pm$ } &  39.7{\scriptsize $\pm$ 1.0} & 70.9{\scriptsize $\pm$ } & 60.0{\scriptsize $\pm$ }\\
       \modelname{}  &\ding{55} &\text{subset}& 89.8{\scriptsize $\pm$1.7} & 85.9{\scriptsize $\pm$0.7} & 80.8{\scriptsize $\pm$1.2} &  \textbf{75.3}{\scriptsize $\pm$0.5} & 82.6{\scriptsize $\pm$1.7} & \textbf{79.9}{\scriptsize $\pm$2.5} & 79.6{\scriptsize $\pm$2.0} & \textbf{68.7}{\scriptsize $\pm$1.1}\\
       \bottomrule
    \end{tabular}
    \caption{Average (AVG) and worst-group accuracy (WGA) on the test set for various shortcut mitigation methods applied on Waterbirds-95, Waterbirds-100 and ISIC. \ding{55} - not using group labels or masked annotations, \ding{51} using group labels during training. We \textbf{bold} the highest WGA; results are averaged across five runs.} 
    \label{tab:pre-result}
\end{threeparttable}
\end{table*}

\subsection{Mitigation of Reliance on Spurious Features} 
\label{ssec:approach:mitigation}
After identifying spurious-positive samples, we propose a regularized fine-tuning procedure that suppresses neuron activations associated with spurious features. To this end, we extract relevance patterns attributable to spurious input regions from the paired masked and original samples $(M^{y_k}, I^{y_k})$. During fine-tuning, we use the masked inputs to isolate neuron activations associated with spurious features. 

\subsubsection{Spurious Contribution Embeddings.} 
For each masked input $m_i$, we define the \emph{spurious relevance score} of neurons at layer $l$ with respect to class $y_k$ as
\begin{equation}
    \mathbf{r}^{l}_i = \mathbf{R}^{l}_{y_k \leftarrow \cdot}(m_i),
\end{equation}
which represents the relevance of neurons in layer $l$ for predicting class $y_k$ when only spurious regions are retained. We further define the \emph{average spurious relevance} across all spurious-positive samples as
\begin{equation}
    \bar{\mathbf{r}}^{l} = \frac{1}{t} \sum_{i=1}^t \mathbf{r}^{l}_i .
\end{equation}

For each paired masked and original sample $(m_i, x_i) \in (M^{y_k}, I^{y_k})$, we define the \textbf{spurious contribution embedding} at layer $l$ as
\begin{equation} \label{eqn:pos-emb}
    \mathbf{z}^{l}_i = f^{l}(x_i) \odot \mathrm{ReLU}\!\left( \mathbf{r}^{l}_i \right),
\end{equation}
where $f^{l}(x_i)$ denotes the activation of sample $x_i$ at layer $l$ of model $f$. The embedding $\mathbf{z}^{l}_i$ isolates neuron activations attributable to spurious input features for sample $i$.
Since the $\epsilon$-rule yields signed relevance, positive relevance corresponds to neuron contributions that support the prediction of class $y_k$ via spurious features, we apply $\mathrm{ReLU}$  to retain only positive relevance scores.

For any other training sample $x_j \in \mathcal{X}$ with label $y_k$, we define the corresponding \textbf{spurious contribution embedding} using the average spurious relevance as
\begin{equation} \label{eqn:avg-emb}
    \bar{\mathbf{z}}^{l}_j = f^{l}(x_j) \odot \mathrm{ReLU}\!\left( \bar{\mathbf{r}}^{l} \right).
\end{equation}

\subsubsection{Regularization.}
Given the spurious contribution embedding of instances in the positive sample set  $\mathbf{z}^{l}_i$ and any other instance $\bar{\mathbf{z}}^{l}_j$, we define a regularization term applied to samples of class $y_k$ as
\begin{equation}\label{eq:reg}
    \mathcal{R}^{y_k} = \sum_{i \in I^{y_k}} \|\mathbf{z}^{l}_i\|_1 \;+\; \sum_{j \notin I^{y_k}} \|\bar{\mathbf{z}}^{l}_j\|_1 ,
\end{equation}
where $\|\cdot\|_1$ denotes the $\ell_1$ norm.
The first term provides a supervised penalty for spurious-positive samples, in which each activation is paired with a corresponding mask that explicitly identifies neuron contributions associated with spurious features via the input pair $(m_i, x_i)$.  
The second term extends this supervision to other samples of class $y_k$ by penalizing neuron activations aligned with the average spurious relevance pattern, thereby targeting neurons that consistently respond to spurious features.
This regularization suppresses large values in the spurious contribution embeddings, which are designed to capture neuron activations attributable to spurious input regions. By constraining these activations, the model is discouraged from relying on shortcut features and is instead biased toward representations that support more robust and generalizable predictions. Without loss of generality, the regularization need not be restricted to a single intermediate layer and can be applied to multiple layers.

\subsubsection{Fine-tuning.} 
For fine-tuning, we construct a dataset $\mathcal{D}_{ft}$ by sampling a subset of the training data and augmenting it with the spurious-positive pairs $(M^{y_k}, I^{y_k})$. 
The objective of fine-tuning is to suppress neuron activations associated with spurious features in the intermediate layer by minimizing the relevance-based regularization term $\mathcal{R}^{y_k}$. 
Since the model is already pretrained, only a small number of fine-tuning epochs is required; in practice, we fine-tune for at most 20 epochs.

The overall batch loss is defined as
\begin{equation} \label{eqn:loss}
    \mathcal{L} = \mathcal{L}^{\mathrm{CE}}(\mathbf{x}, y) + \alpha \mathcal{R}^{y_k},
\end{equation}
where $\mathcal{L}^{\mathrm{CE}}$ denotes the standard cross-entropy loss and $\alpha$ is a weighting hyperparameter.

\section{Results}
\label{sec:results}
We evaluate our approach on both synthetic and real-world datasets that exhibit varying types and strengths of spurious correlations, including single-class and multi-class shortcuts with different degrees of correlation. 
In addition, we analyze the influence of the positive sample set size, regularization strategies, and the availability of ground-truth segmentation masks as an upper bound in Sec. \ref{sec:ablation}. Extension results for Vision Transformers are reported in Sup. \ref{suppl:vit}.

\subsection{Experimental Setup} \label{sec:setup}
\subsubsection{Datasets.} 
We evaluate on four datasets. \textbf{Waterbirds}, built from CUB and Places, requires distinguishing waterbirds from landbirds while the background acts as spurious attribute; we use the benchmark Waterbirds-95 setting and a Waterbirds-100 variant with perfect spurious correlation in train set. \textbf{ISIC} is a real-world skin cancer dataset in which colored patches correlate with the benign class, as no malignant training image contains them. \textbf{Knee Radiographs} targets osteoarthritis classification~\cite{Kuhn:2025:ICCV} contains spurious hospital tag correlated with healthy patients. See Suppl.~\ref{suppl:sec:data} and Tab.~\ref{tab:dataset_overview} for details, and Fig.~\ref{fig:lrp-mask} for examples.

\subsubsection{Baselines.} 
We compare \modelname against state-of-the-art mitigation methods: GroupDRO~\cite{Sagawa:2020b:ICML} and DFR$_{Tr}$~\cite{Kirichenko:2023:ICLR}, which optimize the worst-group loss from training group labels, and the annotation-free JTT~\cite{Liu:2021:ICML}, CnC~\cite{Zhang:2021:ICML}, MaskTune~\cite{Asgari:2022:NeurIPS} and PruSC~\cite{Le:TMLR:2025}. DFR$_V$, SELF~\cite{Labonte:NeurIPS:2024} and NeuronTune~\cite{Zheng:ICML:2025} all require a held-out validation set, we compare against our \modelname$_V$, which uses the same data. CDEP~\cite{Rieger:2020:ICML} shares our \emph{right for the right reason} objective, but suppresses unwanted regions via explanation penalties over pre-specified pixel masks rather than internal representations.

\subsubsection{Metrics.}
We assess the effectiveness of spurious feature mitigation using worst-group accuracy (WGA) and average accuracy (AVG). WGA measures performance on the subgroup with the lowest accuracy, reflecting robustness to spurious correlations, while AVG captures overall predictive performance:
\begin{equation*}
    \mathrm{WGA}(f_{\theta}) = \min_{g \in \mathcal{G}} \; \mathbb{E}_{(x, y) \sim \mathcal{P}_g} \left[ \mathds{1}\!\left(f_{\theta}(x) = y\right) \right],
\end{equation*}
\begin{equation*}
    \mathrm{AVG}(f_{\theta}) = \mathbb{E}_{(x, y) \sim \mathcal{P}_{\mathcal{D}}} \left[ \mathds{1}\!\left(f_{\theta}(x) = y\right) \right],
\end{equation*}
where $\mathds{1}(\cdot)$ denotes the indicator function, $\mathcal{G}$ is the set of groups, and $\mathcal{P}_g$ and $\mathcal{P}_{\mathcal{D}}$ denote the joint data distributions of group $g$ and of the full dataset $\mathcal{D}$, respectively.

\subsubsection{\modelname Parameters and Training.}
For spurious feature detection, we analyze the top-$n_{\text{ref}}=100$ samples with the highest class activation for each class in each dataset.
All experiments are conducted using a ResNet50 backbone.
For fine-tuning \modelname, we construct a class-balanced dataset by sampling additional data from the training set and combining it with the spurious-positive samples $I^{y_k}$. 
Unless otherwise stated, the fine-tuning dataset contains 1000 samples and the model is fine-tuned for 20 epochs. Results under this setting are reported in Tab. \ref{tab:pre-result}. 
For \modelname{}$_{\mathrm{V}}$, we replace the sampled training subset with the validation set and combine it with the positive sample set, yielding a group-balanced held-out dataset comparable to the DFR$_{\mathrm{V}}$ setting.

All fine-tuning variants optimize the loss defined in Eq. \ref{eqn:loss} and apply regularization to the penultimate layer. 
We report the worst-group accuracy achieved with $\alpha = 0.05$. 
Other hyper-parameters follow the settings of GroupDRO~\cite{Sagawa:2020b:ICML} and DFR~\cite{Kirichenko:2023:ICLR}. 
Additional ablations are provided for variants of fine-tuning set size and multi-layer penalization in Suppl. \ref{suppl:ablation}.

\begin{table}[t]
    \centering
    \small
    \setlength{\tabcolsep}{2pt}
    \begin{tabular}{lcccc}
        \toprule
         & WB95 & WB100 & ISIC & Knee \\
        \midrule
        \modelname{}$_{\mathrm{V}}$ & 89.2{\scriptsize $\pm$1.3} & \textbf{90.7}{\scriptsize $\pm$0.9} & \textbf{80.9}{\scriptsize $\pm$0.8} & \textbf{81.0}{\scriptsize $\pm$1.0} \\
        DFR$_{\mathrm{V}}$ & \textbf{91.8}{\scriptsize $\pm$2.6} & 78.8{\scriptsize $\pm$3.7} & 77.7{\scriptsize $\pm$2.4} & 66.9{\scriptsize $\pm$4.1} \\
        NeuronTune & 91.0{\scriptsize $\pm$} & 87.7{\scriptsize $\pm$} & 73.4{\scriptsize $\pm$} & 72.0{\scriptsize $\pm$} \\
        SELF & 92.5{\scriptsize $\pm$} & 32.5{\scriptsize $\pm$} & 68.8{\scriptsize $\pm$} & 48.0{\scriptsize $\pm$} \\
        \bottomrule
    \end{tabular}
\caption{WGA for methods using group-balanced held-out data during training and fine-tuning. We \textbf{bold} the highest WGA; results are averaged across five runs.}
\label{tab:result-w-val}
\end{table}

\subsection{Quantitative Results} 
\label{ssec:results:mitigation}

We report worst-group accuracy (WGA) and average accuracy (AVG) for all methods in Tab. \ref{tab:pre-result}, comparing \modelname with existing spurious-mitigation baselines. Overall, \modelname achieves the strongest performance across three out of four datasets and 4\% gap compared to GroupDRO even GroupDRO using group labels. Results for a ViT backbone are reported in Suppl. F, where \modelname likewise improves WGA on all datasets by a large margin.

\paragraph{Mitigation even without minority groups.}
Unlike prior methods that assume the availability of training data from all groups~\cite{Sagawa:2020a:ICLR,ZhangMichael:2022:ICML}, our approach remains effective in settings where one or more groups are absent from the training data. This is particularly evident on Waterbirds-100 and ISIC, where at least one group is missing. In these extreme, but realistic cases, our method outperforms baselines that rely on training group annotations, such as GroupDRO and DFR$_{\mathrm{Tr}}$ by a large margin.

\paragraph{Recovering under-learned core features.} Methods such as DFR, SELF and NeuronTune assume that the baseline model has already learned all relevant features, so that robustness is a matter of reweighting them. The Knee dataset violates this assumption: the inserted radiographic markers dominate learning so completely that all top-activating healthy samples are explained by the marker alone (Suppl.~\ref{suppl:sec:shortcut-detection}). Reweighting, therefore, has little to reweight: DFR$_\text{Tr}$ falls to 26.0\% WGA, below the 35.3\% baseline, and DFR$_\text{V}$ reaches only 66.9\% even with a group-balanced held-out set (Tab.~\ref{tab:pre-result}, Tab.~\ref{tab:result-w-val}). SCORE instead fine-tunes the entire network under regularization, allowing the model to relearn informative features while being explicitly guided by the ``should not use'' signal from spurious-positive samples. This enables SCORE to reach 68.7\% WGA on Knee, substantially outperforming other methods.

\paragraph{Right for the right reason.} CDEP~\cite{Rieger:2020:ICML} shares our objective of aligning model reasoning with causal features, but requires pixel-level annotation of the spurious region and prior knowledge of which concept is spurious. SCORE requires neither: annotators mark spurious-positive samples directly from LRP heatmaps, with no concept specified in advance. \modelname improves substantially over the CDEP baseline while requiring the least prior specification of spurious features.

\paragraph{Leveraging held-out data.} 
Tab. \ref{tab:result-w-val} compares methods in the (unrealistic) setting where a group-balanced validation set is available. 
Across all datasets, fine-tuning with a group-balanced validation set consistently improves performance over fine-tuning with a subset of the training data (i.e., \modelname{}$_{\mathrm{V}}$ vs.\ \modelname). In Waterbirds-100 and ISIC, the validation set additionally contains samples from groups that are absent in the training data.  Among three out of four datasets (WB100, ISIC, Knee), \modelname{}$_{\mathrm{V}}$ outperforms methods that strictly require a validation set (DFR$_{\mathrm{V}}$, NeuronTune) and two out of four datasets (ISIC, Knee), even SCORE without a validation set outperforms DFR with it.

\subsection{Qualitative Analysis}

\begin{figure*}[ht]
\centering
\includegraphics[width=0.7\textwidth]{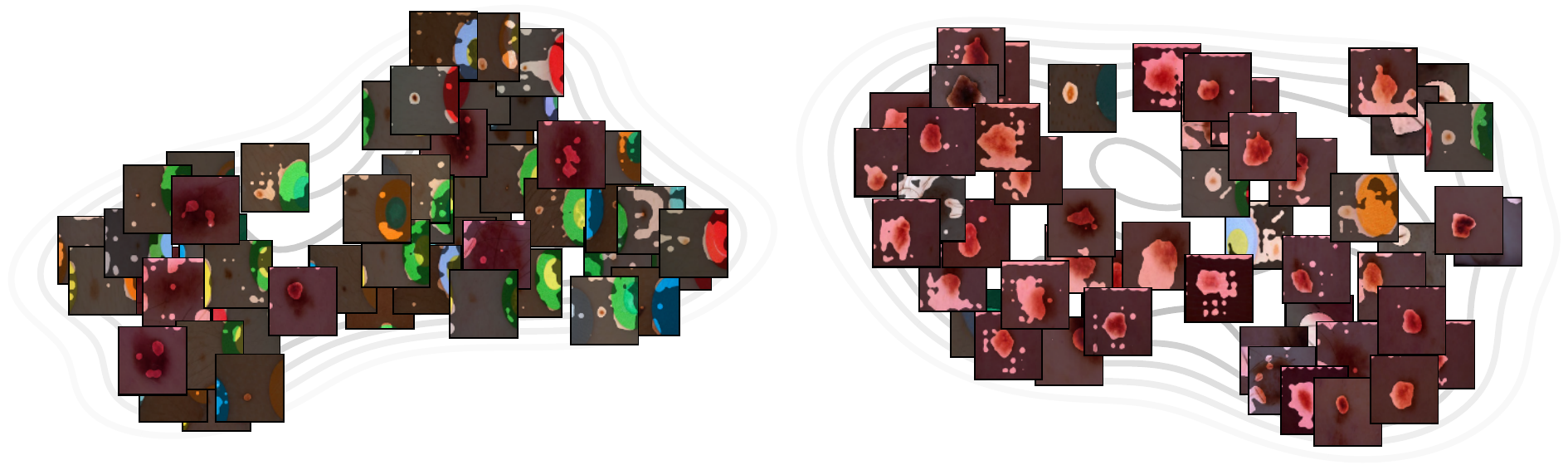}
\caption{UMAP embeddings of the top-50 benign samples, with activating image regions highlighted, before (left) and after (right) fine-tuning. After fine-tuning, the color patches are no longer the most representative features for the target class.}
\label{fig:concept-emb-retrain}
\end{figure*}

We further examine the effect of mitigation qualitatively by analyzing (i) which features remain most relevant for class prediction and (ii) how spurious features are reflected in local explanations.

To assess \emph{class-most-relevant features}, we apply LRP to the most activating samples per class. Fig. \ref{fig:concept-emb-retrain} shows 2D UMAP \cite{umap:2018} embeddings of the top-50 true positive samples of the benign class in ISIC before fine-tuning (left) and after applying our method (right). After \modelname, samples containing color patches are no longer among the most representative examples for the benign class, indicating a reduced reliance on this spurious feature. 
We further compare \emph{local relevance patterns} before and after fine-tuning for individual ISIC samples. As shown in Fig. \ref{fig:preview}, our method consistently decreases attention to spurious regions and even outperforms GroupDRO using full annotation. Additional examples are provided in Suppl. \ref{suppl:qualitative}.

\subsection{Ablation Study} 
\label{sec:ablation}

\begin{table}[t]
    \centering
    \small
    \setlength{\tabcolsep}{4pt}
     \begin{tabular}{llcccc}
    \toprule
         &\multirow{2}{0.7cm}{$|I^{y_k}|$} & \multicolumn{2}{c}{ISIC} & \multicolumn{2}{c}{Knee}\\
          & & AVG & WGA$\uparrow$ & AVG & WGA$\uparrow$ \\
         \midrule
       ResNet50 & 0 & 86.3 & 35.1 & 75.0 & 36.0 \\
       ResNet50* & 0 & 92.5 & 89.0 & 85.0 & 83.0 \\
       \midrule
        \modelname & 25 & 81.5{\scriptsize $\pm$0.8} &  79.4{\scriptsize $\pm$0.6} & 78.9{\scriptsize $\pm$4.0}  & 66.7{\scriptsize $\pm$2.3} \\
        \modelname & 50 & 83.4{\scriptsize $\pm$0.8} &  81.2{\scriptsize $\pm$1.6} & 77.9{\scriptsize $\pm$3.8}  & 68.0{\scriptsize $\pm$2.0} \\
        \modelname & 100 & 83.0{\scriptsize $\pm$0.1} &  80.8{\scriptsize $\pm$1.1} & 80.6{\scriptsize $\pm$2.5}  & 69.0{\scriptsize $\pm$1.4} \\
        \modelname & 200 & 82.8{\scriptsize $\pm$1.3} &  81.0{\scriptsize $\pm$1.0} & 83.4{\scriptsize $\pm$2.0}  & 77.3{\scriptsize $\pm$3.0} \\
        \midrule
        \modelname & 87$^\dagger$ & 83.5 &  77.0 & n.a  & n.a \\
    \bottomrule
    \end{tabular}
    \caption{AVG and WGA of \modelname for varying numbers of positive samples $|I^{y_k}|$, averaged over three runs. ResNet50* trains with spurious-free datasets. $^\dagger$ means $|I^{y_k}|$ containing \emph{mixed} cases instead of truly spurious-positive samples.}
    \label{tab:positive-size}
\end{table}

\subsubsection{Size of Spurious-positive Set.}
Fixing the fine-tuning dataset size to 1,000 samples and varying only the number of positive samples on Knee dataset consistently improves performance, with gains in WGA and AVG when increasing the number of positive samples. Whereas on ISIC, the performance gradually saturates (after 50 positive samples). Relative to a spurious-free ResNet-50 training (Tab. 3, ResNet50*), SCORE closes roughly 85\% of the gap on both datasets.
Additional ablations varying both the fine-tuning dataset size and the positive set size jointly are given in Suppl. D, together with an analysis of noisy selections, where including samples that mix core and spurious evidence, costs 3–7\% of WGA yet still surpasses all baselines.
We also ablate noisy selection$^\dagger$ or \emph{mixed} cases - LRP masks where both core and spurious features are highlighted, on ISIC (among top-100 inspected samples we choose 40 truly spurious-positive and 47 mixed). Mixed cases causes degrading in WGA compare to cleaned ones, however, still surpasses baseline.

\subsubsection{Regularization Terms.}  
\begin{table}[tbp]
    \centering
    \small
    \setlength{\tabcolsep}{3pt}
    \begin{tabular}{lcccc}
        \toprule
        \textbf{Finetuning Set} & $\sum_{i \in I^{y_k}} \|\mathbf{z}^{l}_i\|_1$ & $\sum_{j \notin I^{y_k}} \|\bar{\mathbf{z}}^{l}_j\|_1$ & ISIC & WB100 \\
        \midrule
        $\mathcal{D}_{y\text{-balanced}}$  
            & \ding{55} & \ding{55} & 48.3 &  49.9\\
        $\mathcal{D}_{ft}$      
            & \ding{51} & \ding{55}  & 78.9 &  61.2\\
        $\mathcal{D}_{ft}$ (\modelname) 
            & \ding{51} & \ding{51}  & 81.6 &  76.7\\
        \bottomrule
    \end{tabular}
    \caption{WGA under ablation of the regularization terms, included (\ding{51}) or excluded (\ding{55}). $\mathcal{D}_{y\text{-balanced}}$ is a class-balanced training subset, $\mathcal{D}_{ft}$ SCORE-defined fine-tuning set.
    }
    \label{tab:ablation-reg}
\end{table}

Tab. \ref{tab:ablation-reg} isolates the two terms of Eq.~\ref{eq:reg}. Class-balanced fine-tuning without regularization is insufficient. Penalizing the spurious-positive samples alone improves the WGA (78.9\% and 61.2\%), and extending the penalty to the remaining samples of class $y_k$ via the average relevance vector adds a further 2.7\% and 15.5\% on ISIC and WB100, respectively. The same ordering holds when the group-balanced validation set replaces the training subset (Tab~\ref{tab:ablation-reg-w-val}). 
Extending regularization to additional convolutional layers yields no consistent benefit (Suppl. D).

\subsubsection{LRP Masks vs. Ground-truth Masks.}
\begin{table}[tbp]
    \centering
    \small
    \setlength{\tabcolsep}{3pt}
    \begin{tabular}{lcccc}
        \toprule
         & WB95 & WB100 & ISIC & Knee \\
        \midrule
        LRP masks  & 85.9{\scriptsize $\pm$0.7} & 76.7{\scriptsize $\pm$0.3} & 79.9{\scriptsize $\pm$2.5} & 68.7{\scriptsize $\pm$1.1}\\
        Ground-truth masks  & 86.6{\scriptsize $\pm$1.9} & 78.1{\scriptsize $\pm$1.0} & 75.4{\scriptsize $\pm$1.8} & 72.7{\scriptsize $\pm$2.9} \\
        \bottomrule
    \end{tabular}
    \caption{WGA for \modelname with dynamic LRP masks, vs. ground-truth segmentation mask, averaged over five runs.}
    \label{tab:ground-truth-mask}
\end{table}
If prior knowledge about spurious correlations is available, ground-truth masks can be used to construct the positive sample set $M^{y_k}_{gt}$. This setting serves as an approximate upper bound, as it assumes explicit knowledge of which regions constitute spurious features. 
On the synthetic benchmarks, the two settings are close (the gaps are 1.5\% on Waterbirds; ground truth outperforms by 4\% on Knee). On the real-world ISIC data, LRP masks outperform ground truth, since they additionally capture secondary cues such as hair and ruler marks (Fig.~\ref{fig:lrp-mask}) that are i) unknown a priori and ii) impractical to segment.

\section{Conclusion} \label{sec:conclusion}
In this paper, we highlighted that common requirements of shortcut mitigation, training data annotations, group information or the existence of all groups in training data, are rarely met in practice, and that benchmarks hide these challenges.
We proposed a method that instead relies on a small set of spurious-positive samples, collected with little effort during model inspection, and showed that shortcut learning can be effectively mitigated without any of these requirements. We encourage the community to focus research on realistic settings to develop robust models that work in practice.

\bibliography{mylib}

\appendix

\section{Datasets} \label{suppl:sec:data}
\begin{table}[t]
\centering
\small
\setlength{\tabcolsep}{6pt}
\begin{tabular}{lccccc}
\toprule
\textbf{Dataset} & \textbf{Class} & $p(y)$ & \textbf{Spurious attr.} & $p(s{=}1\mid y)$ \\
\midrule
\multirow{2}{*}{WB95}  
 & Waterbird & 0.22 & Water & 0.95 \\
 & Landbird & 0.77 & Land  & 0.95 \\
\midrule
\multirow{2}{*}{WB100} 
 & Waterbird & 0.23 & Water & 1.00 \\
 & Landbird & 0.77 & Land  & 1.00 \\
\midrule
ISIC      
 & Benign & 0.88 & Patch & 0.47 \\
 & Malignant & 0.12 & Patch & 0.00 \\
\midrule
\multirow{2}{*}{Knee}     
 & Normal & 0.50 & RM & 0.50 \\
 & Abnormal & 0.50 & RM & 0.03 \\
\bottomrule
\end{tabular}
\caption{Dataset overview. Spurious attribute $s$: background (Water/Land) for Waterbirds, patch (P) for ISIC, radiographic marker (RM) for Knee.}
\label{tab:dataset_overview}
\end{table}
\paragraph{ISIC.} ISIC, provided by the International Skin Imaging Collaboration ISIC, is a real-world skin cancer detection dataset. Following the instruction from~\cite{Rieger:2020:ICML}, we retrieved \(20, 394\) images, including 17,881 \texttt{benign} and 2,513 \texttt{malignant}. Nearly half of the \texttt{benign} images contain colored patches, while none of the \texttt{malignant} samples do. 
Therefore, the dataset contains an extreme case of spurious correlation where one group is entirely absent from the training set (malignant with patches). We artificially insert 821 test images and 60 validation images of the group malignant with patches.
\paragraph{Waterbirds-x.} The Waterbirds (WB) dataset~\cite{Sagawa:2020a:ICLR} is synthetic and constructed from the CUB~\cite{Wah:2011:cub-dataset} and Places~\cite{Zhou:2018:place-dataset} datasets. The classification task is to distinguish between \texttt{waterbirds} and \texttt{landbirds}, but the dataset is designed with strong spurious correlations: bird species are strongly correlated with the background. The strength of spurious correlation is defined by a pre-fixed \textbf{x}\% during the generation process. The standard benchmark, Waterbirds-95, contains 95\% spurious correlation in the training set, meaning that 95\% of waterbirds have water backgrounds and 95\% of landbirds on land backgrounds. In our experiments, we additionally use the WB100 variant, in which the training set exhibits 100\% spurious correlation, but the test set remains the same as that used for WB95.
\paragraph{Knee Radiographs.} The knee osteoarthritis radiograph dataset~\cite{knee:2019} consists of knee joint X-ray images from 4,796 participants exhibiting varying degrees of osteoarthritis. We address the challenging binary classification task of distinguishing between `no' versus `moderate' osteoarthritis. Following prior work~\cite{Kuhn:2025:ICCV}, we introduce a spurious hospital tag to 50\% of the healthy images and 2.5\% of the arthritic images. The training contains 1,000 samples per class. For each dataset of validation and testing, we generate 100 samples per class, both with and without the spurious marker (50\%). The worst group in this dataset corresponds to moderate arthritis images containing the added radiographic marker.

\section{Shortcut Detection in Real-world Dataset} \label{suppl:sec:shortcut-detection}

\begin{figure*}[h]
\begin{center}
\includegraphics[width=0.45\textwidth,trim={1.5cm 1.5cm 1.5cm 0.3cm},clip]{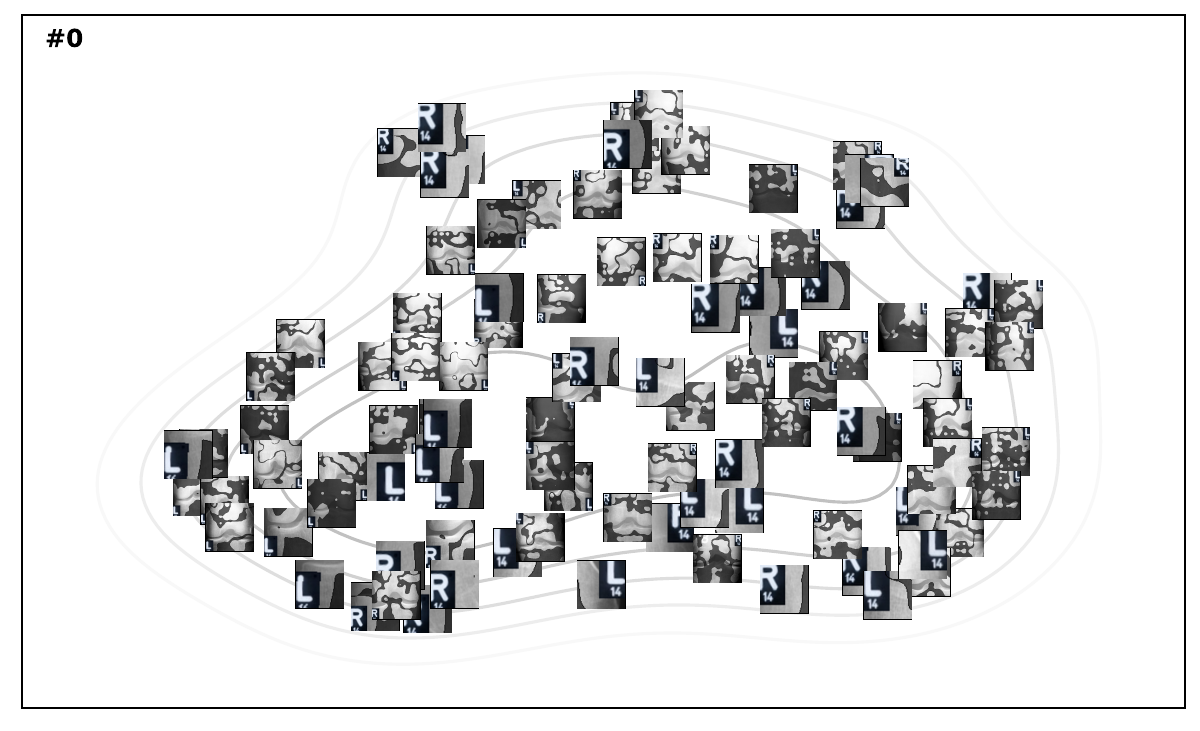}
\includegraphics[width=0.45\textwidth,trim={0.5cm 5.5cm 6.5cm 0.3cm},clip]{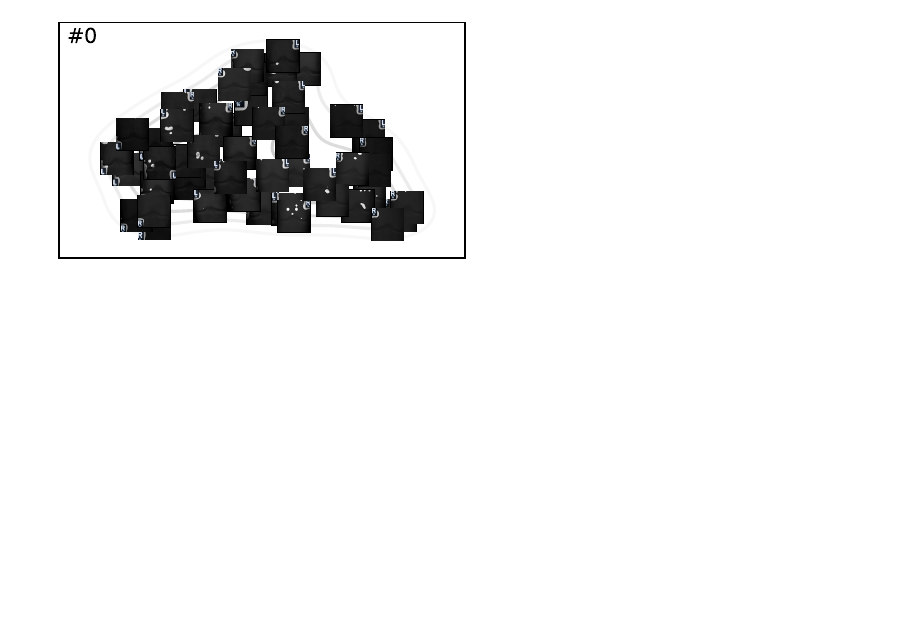}
\end{center}
\caption{Left: 2D-UMAP visualization of the Knee dataset showing the most relevant image patches for class predictions. Many highlighted regions correspond to artificially inserted patches, indicating the model’s reliance on spurious features. Right: 2D-UMAP visualization shows the most relevant image patches for the healthy class; lighter regions are more important.}
\label{fig:concept-emb-knee}
\end{figure*}

\paragraph{Knee Radiographs dataset.} Considering the whole dataset, Fig. \ref{fig:concept-emb-knee} highlights the regions most strongly associated with model predictions. Upon inspection, humans can potentially identify the spurious patches. Indeed, due to the way the dataset was constructed, these patches correspond to artificially inserted features that were assigned to the healthy class. Notably, as shown in Fig. \ref{fig:concept-emb-knee} (right), all of the top 100\% most activating samples for the healthy class highlight the artificially inserted patches as the most relevant regions for the prediction. This indicates that the model relies almost entirely on these artificial cues when making predictions for this class, suggesting that it has learned the spurious correlation rather than the underlying semantic features of the knee images. Such behavior is unexpected and may highlight a limitation of the synthetic dataset, where the artificially inserted patches become overly dominant compared to the original image features, leading the model to prioritize them during learning.

\begin{figure*}[h]
\begin{center}
\includegraphics[width=0.8\textwidth,trim={0cm 6.2cm 0cm 0cm},clip]{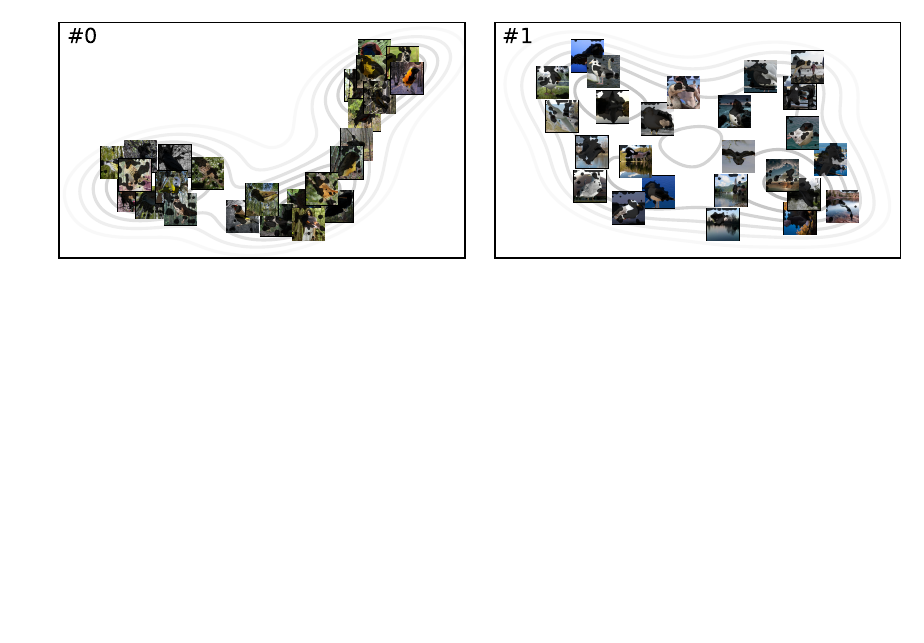}
\end{center}
\caption{2D-UMAP visualization of the top-$25$ samples strongly activate the prediction of the class landbird (left) and waterbird (right) from the dataset Waterbird-100.}
\label{fig:class-emb-wb100}
\end{figure*}

\paragraph{Waterbird100 dataset.} Fig. \ref{fig:class-emb-wb100} shows a UMAP visualization of the 25 samples most representative of the two classes in the Waterbirds dataset. The visualization reveals distinct bias patterns between the classes: landbirds are often associated with bamboo forest backgrounds, while waterbirds frequently appear with water or blue-sky backgrounds. These patterns allow potential spurious features for each class to be identified. However, manual inspection is still necessary to ensure the quality and correctness of the selected spurious-positive samples. 
This highlights the importance of human-in-the-loop intervention for accurately detecting spurious correlations, requiring effort only comparable to model analysis. 

\begin{figure*}[h]
\begin{center}
\includegraphics[width=0.4\textwidth,trim={0cm 0cm 0cm 0cm},clip]{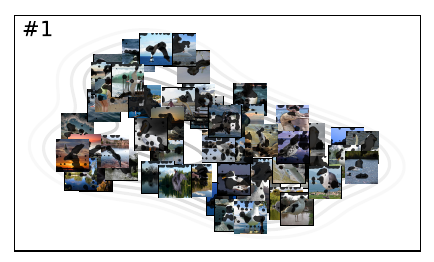}
\includegraphics[width=0.4\textwidth,trim={2.0cm 2.0cm 2.0cm 2.0cm},clip]{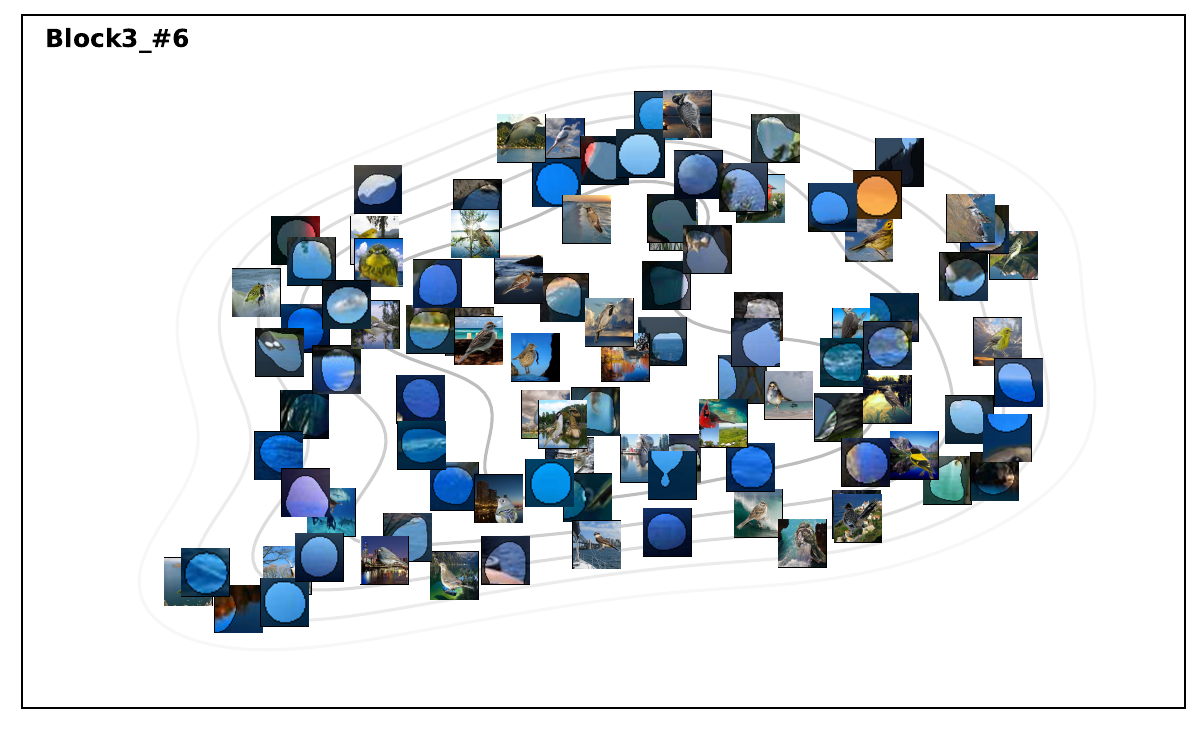}
\end{center}
\caption{ Top-$50$ samples strongly activate the prediction of class waterbird (left) from the dataset WB100. Semantic meanings of spurious neurons (right).}
\label{fig:spurious-emb-wb100}
\end{figure*}

To further investigate, we examine the highly activated neurons based on the masked inputs $M_{y_k}$ to analyze the semantic meanings of these neurons. The representative concepts for this neuron are visualized in Fig.~\ref{fig:spurious-emb-wb100}--right. Interestingly, the neuron’s concept reveals a strong emphasis on the `blue water surface', a spurious feature that is difficult to detect in the initial visualization.

\begin{figure*}[h]
\centering
\includegraphics[width=0.4\textwidth,trim={0cm 0.15cm 7.5cm 0.2cm},clip]{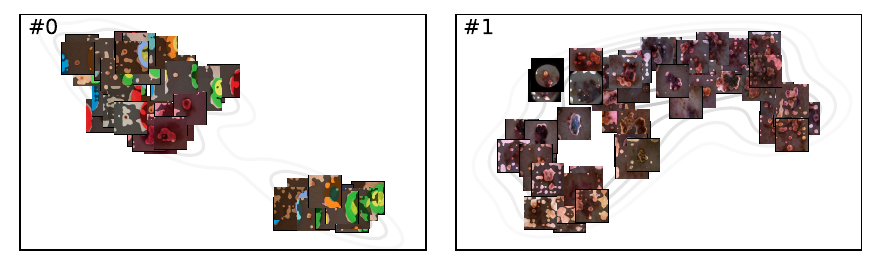}
\includegraphics[width=0.4\textwidth,trim={0cm 0.1cm 7.5cm 0cm},clip]{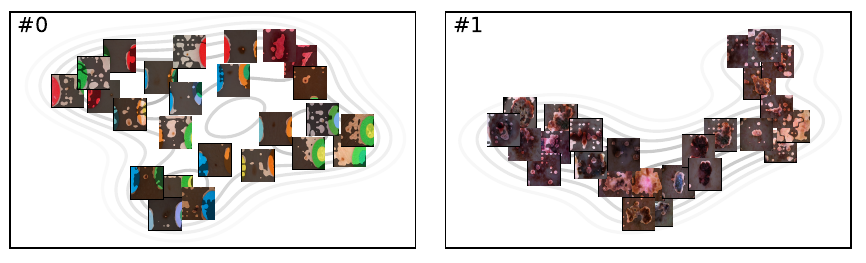}

\caption{Top-$m$ samples strongly activate the prediction of the class benign in ISIC dataset. We select $m = 50$ (left) and $m=25$ (right). We can observe the spurious features (patches) easily, even in the top few samples activating the class.}
\label{fig:concept-emb-isic}
\end{figure*}

\begin{figure}[h]
\centering
\includegraphics[width=0.21\textwidth,trim={0cm 0.2cm 0cm 0.1cm},clip]{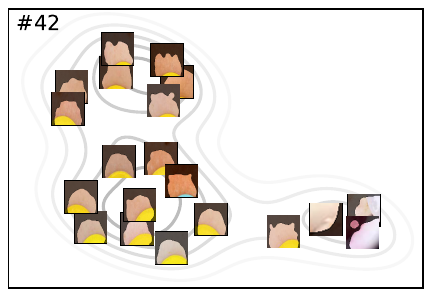}
\includegraphics[width=0.21\textwidth,trim={0cm 0.2cm 0cm 0.1cm},clip]{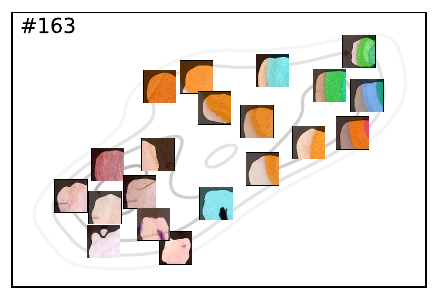}
\includegraphics[width=0.21\textwidth,trim={0cm 0.2cm 0cm 0.1cm},clip]{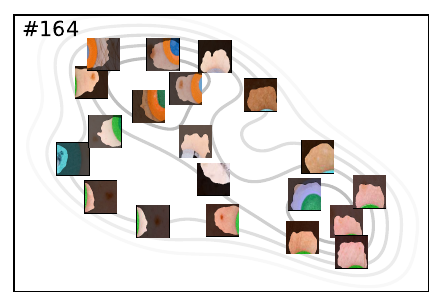}
\includegraphics[width=0.21\textwidth,trim={0cm 0.2cm 0cm 0.1cm},clip]{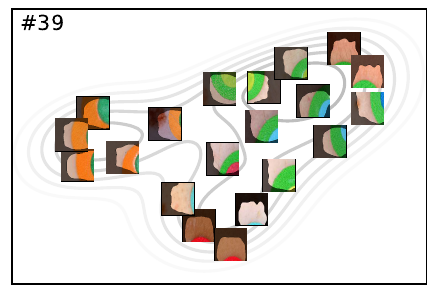}
\caption{2D UMAP embedding of neurons strongly activating for spurious features.}
\label{fig:spurious-emb-isic}
\end{figure}

\paragraph{ISIC Dataset.}
For ISIC dataset, we analyze several cases of top-$m$ samples activating the benign class. Fig. \ref{fig:concept-emb-isic} shows examples for top-25 and top-50 samples activating class benign. We observe that even within only 25 samples, both core features and spurious features (patches) are detected, showing the effectiveness of LRP to decompose important features of a class. Further down, in Fig. \ref{fig:spurious-emb-isic} we show examples of top spurious neurons.

\section{Further Qualitative Analysis} \label{suppl:qualitative}
\begin{figure*}[ht]
\centering
\includegraphics[width=0.8\textwidth]{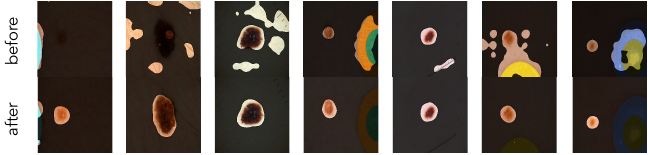}
\caption{LRP heatmaps for samples from the baseline model (top) and after fine-tuning using \modelname (bottom). The results show that the proposed regularization effectively eliminates the reliance on spurious features (color patches, rulers, hairs).}
\label{fig:local-xai}
\end{figure*}

We further compare local relevance patterns before and after fine-tuning for individual ISIC samples. As shown in Fig. \ref{fig:local-xai}, our method consistently decreases attention to spurious regions relative to the baseline. In particular, the model’s focus shifts from purely spurious features to core lesion features (first two columns), and from mixed spurious and core features to core features only (third to last row). In addition, other spurious cues, such as hair and ruler marks, are effectively suppressed (third and fifth columns).

\section{Further Ablation Study} \label{suppl:ablation}
\subsection{Fine-tuning Dataset Construction}
\label{suppl:ablation:datasize}
We analyzed the effect of the number of positive samples and the number of total samples in the fine-tuning data set. 
We denote the set of positive images as $\mathcal{I}^{y_k}$. To construct the fine-tuning dataset, we retain the positive set and sample additional data points from the training set to create a class-balanced dataset, denoted as $\mathcal{D}_{ft}$. Let $|\mathcal{D}_{ft}|$ be the total number of samples in $\mathcal{D}_{ft}$, and $|\mathcal{I}^{y_k}|$ the number of positive samples. 

Given $K$ the number of classes in the dataset, pre-defined the size of fine-tuning set $|\mathcal{D}_{ft}|$, and $\mathcal{I}^{y_k}$ the positive sample set from class $y_k$, if there is no spurious correlation detected w.r.t class $y_k$ then $|\mathcal{I}^{y_k}| = 0$, i.e., the set is empty, the number of data points having label $y_k$ we need to draw from training set is:
$$ \frac{|\mathcal{D}_{ft}|}{K} - |\mathcal{I}^{y_k}| $$

For each training iteration, the loss is computed as
\begin{equation*}
    \mathcal{L} = \mathcal{L}^{\mathrm{CE}} + \alpha \left( \sum_{i \in \mathcal{I}^{y_k}} \|\mathbf{z}^{l}_i \|_1 + \sum_{j \notin \mathcal{I}^{y_k}} \| \mathbf{\Bar{z}}^{l}_j \|_1 \right)
\end{equation*}
where $\mathbf{z}$ and $\mathbf{\Bar{z}}$ are defined in Eq. (5) and Eq. (6) respectively. The average relevance vector $\mathbf{\Bar{z}}$ is computed by averaging all relevance vectors $\mathbf{z}$ from samples in $\mathcal{I}^{y_k}$ and is re-computed at each iteration.

Suppl. \ref{suppl:tab:datasize} reports the experiments on the ISIC and Knee datasets conducted with different sizes of the fine-tuning dataset ($|\mathcal{D}_{ft}|$) and varying numbers of positive samples ($|\mathcal{I}^{y_k}|$). For the number of positive samples, we ablate the values at 25, 50, 100, and 200. For each case, we investigate fine-tuning datasets of sizes 300, 500, and 1000, sampled in a class-balanced manner from the training set. Overall, the empirical results indicate that the optimal choice of the number of positive samples and fine-tuning dataset size is dataset-dependent. For the Knee dataset, the best performance is achieved with 200 positive samples and 1000 fine-tuning samples, whereas for ISIC, strong performance is obtained with 100 positive samples and 500 fine-tuning samples.

Across both datasets, worst-group accuracy (WGA) consistently improves as the size of the fine-tuning set $\mathcal{D}_{ft}$ increases. This trend is consistent with the effect of the average spurious-relevance regularization, as more samples are subject to the penalty during fine-tuning. Increasing the number of positive samples leads to moderate performance gains, with saturation observed for ISIC beyond 100 positive samples. Notably, competitive performance can already be achieved with as few as 50 positive samples.

\begin{table}[tbhp]
    \centering
    \setlength{\tabcolsep}{5pt}
     \begin{tabular}{llcccc}
    \toprule
         \multirow{2}{0.7cm}{$|I^{y_k}|$}&\multirow{2}{0.7cm}{$|\mathcal{D}_{ft}|$} & \multicolumn{2}{c}{ISIC} & \multicolumn{2}{c}{Knee}\\
          & & AVG & WGA$\uparrow$ & AVG & WGA$\uparrow$ \\
         \midrule
        25 & 300 & 81.2{\scriptsize $\pm$0.8} &  75.9{\scriptsize $\pm$1.3} & 76.8{\scriptsize $\pm$3.9}  & 61.0{\scriptsize $\pm$1.4} \\
        25 & 500 & 80.8{\scriptsize $\pm$1.2} &  77.3{\scriptsize $\pm$2.8} & 80.0{\scriptsize $\pm$0.3}  & 64.7{\scriptsize $\pm$3.0} \\
        25 & 1000 & 81.5{\scriptsize $\pm$0.8} &  79.4{\scriptsize $\pm$0.6} & 78.9{\scriptsize $\pm$4.0}  & 66.7{\scriptsize $\pm$2.3} \\
        \midrule
        50 & 300 & 82.6{\scriptsize $\pm$1.7} &  76.9{\scriptsize $\pm$2.0} & 78.0{\scriptsize $\pm$3.7}  & 65.3{\scriptsize $\pm$2.3} \\
        50 & 500 & 82.7{\scriptsize $\pm$0.4} &  78.6{\scriptsize $\pm$2.1} & 77.9{\scriptsize $\pm$3.8}  & 66.7{\scriptsize $\pm$2.3} \\
        50 & 1000 & 83.4{\scriptsize $\pm$0.8} &  81.2{\scriptsize $\pm$1.6} & 77.9{\scriptsize $\pm$3.8}  & 68.0{\scriptsize $\pm$2.0} \\
        \midrule
        100 & 300 & 81.2{\scriptsize $\pm$0.1} &  78.0{\scriptsize $\pm$0.1} & 77.8{\scriptsize $\pm$2.0}  & 64.0{\scriptsize $\pm$0.0}\\
        100 & 500 & 82.9{\scriptsize $\pm$0.6} &  81.5{\scriptsize $\pm$0.2} & 79.2{\scriptsize $\pm$2.0}  & 70.0{\scriptsize $\pm$2.8} \\
        100 & 1000 & 83.0{\scriptsize $\pm$0.1} &  80.8{\scriptsize $\pm$1.1} & 80.6{\scriptsize $\pm$2.5}  & 69.0{\scriptsize $\pm$1.4} \\
        \midrule
        200 & 1000 & 82.8{\scriptsize $\pm$1.3} &  81.0{\scriptsize $\pm$1.0} & 83.4{\scriptsize $\pm$2.0}  & 77.3{\scriptsize $\pm$3.0} \\
    \bottomrule
    \end{tabular}
    \caption{AVG and WGA of \modelname on ISIC and Knee datasets for varying numbers of positive samples $|I^{y_k}|$ and fine-tuning dataset size $|\mathcal{D}_{ft}|$, averaged over three runs.}
    \label{suppl:tab:datasize}
\end{table}

\subsection{Effect of different layer penalties} 
\label{suppl:ablation:layers}
Following prior work~\cite{Kirichenko:2023:ICLR,Le:TMLR:2025,Le:2025:xai} which highlight the importance of intervening in later layers where spurious features are encoded, we report our main results using regularization applied to the penultimate layer. However, as discussed in Sec. 3.2, the regularization can be applied to multiple intermediate layers. We additionally experiment with applying regularization to the embeddings of the last convolutional layer and the last two convolutional layers (Tab. \ref{tab:layer-penalty}. The results show no improvement on ISIC when regularizing multiple layers, while a slight improvement is observed on Waterbirds-100. We hypothesize that the effectiveness of the regularization depends on the specific mechanisms by which spurious correlations are learned in different models and datasets. Consequently, identifying the most effective regularization strategy in real-world applications requires more thorough analysis.

\begin{table}[th]
    \centering
    \setlength{\tabcolsep}{5pt}
    \begin{tabular}{lcc}
        \toprule
        \textbf{Layers} & ISIC & WB100 \\
        \midrule
        Last conv layer  & 79.2 & 75.9 \\
        Last 2 conv layers  & 72.5 & 77.1 \\
        \bottomrule
    \end{tabular}
    \caption{WGA on ISIC and Waterbirds-100 under regularization applied to different numbers of layers.}
    \label{tab:layer-penalty}
\end{table}

\subsection{Effect of noisy selections}
We ablate noisy selection or \emph{mixed} cases, where both core and spurious features are highlighted, on ISIC (Tab. \ref{tab:ablation-noisy}), comparing the use of spurious-only samples (40 samples) against combining spurious-only and mixed samples (87 samples). Including mixed cases decreases WGA by 3-7\%, yet still surpasses baselines.

\begin{table}[h]
\centering
\setlength{\tabcolsep}{4pt}
    \begin{tabular}{lcccc}
    \toprule
    & \multicolumn{3}{c}{\emph{spurious-only}} & \emph{mixed} \\
      & WB95 & Knee & ISIC & ISIC\\
     \midrule
      ViT-B/16  & 0.66 & 0.32 & 0.18 & 0.18\\
      SCORE (ViT)   & \textbf{0.85} &  \textbf{0.65}  & \textbf{0.74} & 0.67\\
      \midrule
      ResNet50 & 0.75 & 0.35  & 0.34 & 0.34\\
      SCORE (RN) & \textbf{0.86} & \textbf{0.69}  & \textbf{0.80} & 0.77\\
      \bottomrule
\end{tabular}
\caption{Worst-group accuracy (WGA).}
\label{tab:ablation-noisy}
\end{table}

\subsection{Effect of Regularization during Fine-tuning}
Tab.~\ref{tab:ablation-reg-w-val} presents the ablation study on the effects of the supervised and average regularization terms for the SCORE scenario, considering both fine-tuning using only the training data and fine-tuning using a group-balanced validation set. The results show that combining both regularization terms yields the best performance in all cases. 

\begin{table}[tbp]
    \centering
    \small
    \setlength{\tabcolsep}{3pt}
    \begin{tabular}{lcccc}
        \toprule
        \textbf{Finetuning Set} & $\sum_{i \in I^{y_k}} \|\mathbf{z}^{l}_i\|_1$ & $\sum_{j \notin I^{y_k}} \|\bar{\mathbf{z}}^{l}_j\|_1$ & ISIC & WB100 \\
        \midrule
        $\mathcal{D}_{y\text{-balanced}}$  
            & \ding{55} & \ding{55} & 48.3 &  49.9\\
        $\mathcal{D}_{ft}$      
            & \ding{51} & \ding{55}  & 78.9 &  61.2\\
        $\mathcal{D}_{ft}$ (\modelname) 
            & \ding{51} & \ding{51}  & 81.6 &  76.7\\
        \midrule
        $\mathcal{V}$ 
            & \ding{55} & \ding{55} & 63.2 & 82.5\\
        $\mathcal{V}_{ft}$ 
            & \ding{51} & \ding{55} & 77.6 &  88.1\\
        $\mathcal{V}_{ft}$ (\modelname{}$_{V}$) 
            & \ding{51} & \ding{51} &  81.7 &  91.1\\
        \bottomrule
    \end{tabular}
    \caption{Ablation of regularization terms, either included (\ding{51}) or excluded (\ding{55}). WGA for models fine-tuned under different training settings. $I^{y_k}$ denotes the spurious-positive sample set, $\mathcal{D}{y\text{-balanced}}$ a class-balanced training subset, $\mathcal{D}_{ft}$ our fine-tuning dataset including 1,000 class-balanced samples from training set, and $\mathcal{V}$ the group-balanced validation set.}
    \label{tab:ablation-reg-w-val}
\end{table}

\section{Baselines Details} \label{suppl:baselines}
SCORE uses LRP on the top-$k$ activating samples per class to identify spurious features that models over-rely on as non-causal features from a human perspective. That is, by design, there are no false positives. 
Crucially, spurious correlations cannot be exhaustively specified beforehand, as they are inherently context- and dataset-dependent.
We do not require prior knowledge of spurious features or failure cases, as annotators identify unwanted features directly from LRP heatmaps.

\paragraph{MaskTune Baseline.} 
Methods such as \textbf{MaskTune} detect spurious correlations by masking important features (via xGradCAM) and forcing the model to learn alternative ones. However, this may introduce new biases, as core features can also be masked without any control. Moreover, identifying spurious features with attribution methods requires threshold tuning, highlighting the need for human-guided analysis. 
While MaskTune reports a WGA of 86.4 on a cleaned variant of WB95, SCORE outperforms MaskTune by far on the standard WB95 benchmark, Knee and ISIC (Tab.~\ref{tab:pre-result}), and also provides greater transparency into which features are considered spurious.

\paragraph{DFR Baseline.} Deep Feature Reweighting (DFR)~\cite{Kirichenko:2023:ICLR} is a state-of-the-art method for robustness to spurious correlations that retrains only the final linear layer of a pretrained model. DFR requires a small group-balanced dataset for retraining. We evaluate two variants of DFR: one in which the small dataset is sampled from the training set (DFR$_{\mathrm{Tr}}$) and one in which it is sampled from the validation set (DFR$_{\mathrm{V}}$), in which group annotations are available.
In the DFR$_{\mathrm{Tr}}$ setting, when certain groups are absent from the training set, we construct the small dataset using only the available groups. For ISIC, we select 120 malignant samples without patches and 60 samples from each benign subgroup, yielding a dataset that is balanced both across classes and with respect to spurious features within the benign class. In the DFR$_{\mathrm{V}}$ setting, we artificially generate missing group data, which applies to ISIC and Waterbirds-100.

\paragraph{PruSC Baseline.} We report results for Waterbirds-95 and ISIC from~\citet{Le:TMLR:2025}. For Waterbirds-100 and Knee, we prune up to 10\% and 30\% of weights, respectively; higher pruning leads to model collapse ($0.0$ WGA).
\section{Vision Transformers} \label{suppl:vit}
\begin{figure}[b]
  \centering
  \includegraphics[width=0.2\linewidth]{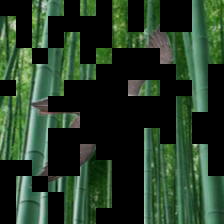}
  \includegraphics[width=0.2\linewidth]{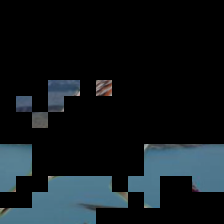}
  \includegraphics[width=0.2\linewidth]{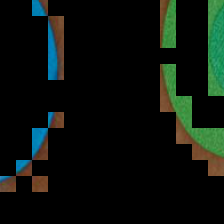}
  \includegraphics[width=0.2\linewidth]{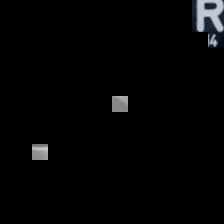}
   \caption{Detected spurious features (\emph{left-to-right}): land background (WB95), water background (WB95), patches (ISIC), markers (Knee).}
   \label{fig:vit-detection}
\end{figure}

Extending SCORE to ViT, we replace the original LRP by AttnLRP~\cite{Achibat:ICML:2024}. Qualitative results show that Attn-LRP successfully highlights class-associated spurious features (Fig. \ref{fig:vit-detection}). We evaluate mitigation performance with 50 human-selected spurious-positive samples for each dataset. The results in Tab.~\ref{tab:vit-result} show that SCORE generalizes beyond CNNs and can be effectively applied to ViT.

\begin{table}[bh]
\centering
\label{tab:vit-result}
\setlength{\tabcolsep}{10pt}
    \begin{tabular}{lccc}
    \toprule
      & WB95 & Knee & ISIC \\
     \midrule
      ViT-B/16  & 0.66 & 0.32 & 0.18 \\
      DFR$_{Tr}$ (ViT)   & 0.73 & 0.52 & 0.30 \\
      SCORE (ViT)   & \textbf{0.85} &  \textbf{0.65}  & \textbf{0.74}\\
      \bottomrule
\end{tabular}
\caption{Worst-group accuracy (WGA).}
\end{table}


\end{document}